\documentclass{article}

\usepackage[preprint]{neurips_2026}

\usepackage[utf8]{inputenc} 
\usepackage[T1]{fontenc}    
\usepackage{hyperref}       
\usepackage{url}            
\usepackage{booktabs}       
\usepackage{amsfonts}       
\usepackage{nicefrac}       
\usepackage{microtype}      
\usepackage{xcolor}         
\usepackage{graphicx}
\usepackage{comment}
\usepackage{subcaption}
\usepackage{algorithm}
\usepackage{algpseudocode}
\usepackage[many]{tcolorbox}

\newtcbtheorem{keyfinding}{Prompt}{
    colback=blue!5,
    colframe=blue!75!black,
    fonttitle=\bfseries
}{kf}

\title{Open-Ended CT Volume Segmentation with Weak Supervision from Language}

\author{
  Sanjay Subramanian\textsuperscript{1}\footnotemark[2] \quad 
  Junwei Yu\textsuperscript{2} \quad 
  Zirui Wang\textsuperscript{1}\footnotemark[2] \quad 
  Rohil Malpani\textsuperscript{2,3} \\
  \bfseries
  Maggie Chung\textsuperscript{2,3} \quad 
  Adam Yala\textsuperscript{2,4} \quad
  Dan Klein\textsuperscript{1} \quad 
  Trevor Darrell\textsuperscript{1}\footnotemark[2] \\
  \vspace{0.2cm}
  \texttt{sanjayss@berkeley.edu} \\
  \textsuperscript{1}Department of Electrical Engineering and Computer Science, UC Berkeley \\ 
  \textsuperscript{2}Voio \\
  \textsuperscript{3}Department of Radiology and Biomedical Imaging, UC San Francisco \\
  \textsuperscript{4}Computational Precision Health, UC Berkeley and UC San Francisco
}

\begin{document}

\maketitle
\footnotetext[2]{Work done at Voio}

\begin{abstract}
  We introduce a method for training a text-conditioned segmentation model for CT scans, which combines voxel-level supervision with coarse but scalable slice-level supervision from reports. We extract, from a large database of scan-report pairs, descriptions of findings with indices of slices where those findings occur. We then finetune a general-purpose 2D image segmentation model, SAM3, with standard segmentation losses from strongly labeled data and with a slice-level classification loss from the extracted weak supervision. Our results on the ReXGroundingCT dataset illustrate that this strategy improves the segmentation dice score: from an $8\%$ relative gain when there are $1000$ fully labeled volumes to $22\%$ when there are $250$ fully labeled volumes.
\end{abstract}

\section{Introduction}
\label{sec:intro}
Segmentation of computed tomography (CT) volumes is challenging because (1) the volumes are large and individual findings are comparatively small \citep{shin2023improving},
(2) there are numerous types of findings that vary widely in shape and size (and variation within a single type, described in natural language) \citep{zhang2025detection}, and (3) ground-truth segmentation data is scarce and requires deep expertise to create \citep{rajpurkar2022ai, wang2024comprehensive}. At the same time, automated CT segmentation could improve tools for radiologists by enabling faster measurement of regions and faster localization of previously annotated findings \citep{pan2024artificial, yacoub2022impact, lenchik2019automated}.

While data with full ground-truth segmentations (i.e. \textit{strong} supervision) is challenging to acquire, CT volumes paired with reports are widely available. Reports enumerating and describing the findings in CT volumes are routinely written by radiologists and saved by institutions. The finding descriptions in these reports reflect the text inputs that would occur in the aforementioned applications of segmentation. In addition, these descriptions often provide indices of slices on which the described finding is prominently visible \citep{wheretolook}. Thus, this data provides a useful \textit{weak} signal for the segmentation task--a (non-exhaustive) indicator of where along the depth axis the finding occurs.

In this work, we introduce an open-ended (i.e. text-conditioned) CT segmentation method that leverages weak supervision from volume-report pairs in addition to strong supervision from ground-truth segmentation data. Specifically, we finetune a general-purpose segmentation model--SAM3 \citep{sam3}--on 2D axial CT slices with both the (strong) dense supervision from ground-truth segmentation data and the (weak) slice-level supervision from report snippets. Compared with finetuning on only strong supervision, our method combining strong and weak supervision achieves an $8\%$ relative gain when there are $1000$ fully labeled volumes and a $22\%$ relative gain when there are $250$ fully labeled volumes. Our work suggests that the challenges of expertise and cost that are associated with scaling supervised training for CT segmentation can be partially mitigated by widely available weak supervision from reports.
\section{Related Work}
\paragraph{Open-ended CT Segmentation} There are a number of approaches which have been proposed for open-ended CT segmentation.\footnote{We use the terms ``open-ended'', ``text-conditioned'', and ``language-conditioned'' interchangeably.} Several of them operate on individual 2D slices--BiomedParse \citep{biomedparse}, MedSAM3 \citep{medsam3}, Medical SAM3 \citep{medicalsam3}. Another line of work has developed open-ended methods which operate on the 3D volume or a subset of slices, such as BiomedParse v2 \citep{biomedparsev2}, SAT \citep{sat}, SegVol \citep{segvol}, Text3DSAM \citep{text3dsam}, and VoxTell \citep{voxtell}. We adopt a per-slice architecture, and 
we are the first to show that weak supervision from reports improves open-ended segmentation performance.

\paragraph{Weak supervision for CT Segmentation}
Prior work has also used the reports paired with CT scans as training data for localization models. For instance, \citet{rsuper} uses measurements in reports as supervision but focuses specifically on tumor segmentation. \citet{wheretolook} uses the slice indices in reports as supervision but trains a model for slice-level localization/image-text matching, not mask/box prediction.  In contrast to this prior work, we use the slice-level signal from reports as weak supervision in the task of \textit{open-ended segmentation}, where the input is a finding description and a volume and the output is a per-voxel segmentation of the volume for that finding. RECIST markers are another source of supervision for CT segmentation \citep{recist,deeplesion}, though they are not widely available across finding types.

\paragraph{Weak supervision for image segmentation}
Our work draws inspiration from the computer vision literature on weak supervision for image segmentation. Research in this direction has developed methods for training segmentation models for natural images using image-level labels or image captions. Some methods are designed to work with only weak supervision \citep{ficklenet,chang2020weakly,ahnWeak,groupvit}, while others combine strong and weak sources of supervision \citep{pseudoseg,decoupled,yu2025unsamv2}.
\section{Volume Segmentation with Weak Supervision}
\label{sec:lang_seg}
\begin{figure*}[ht!]
    \includegraphics[width=\textwidth]{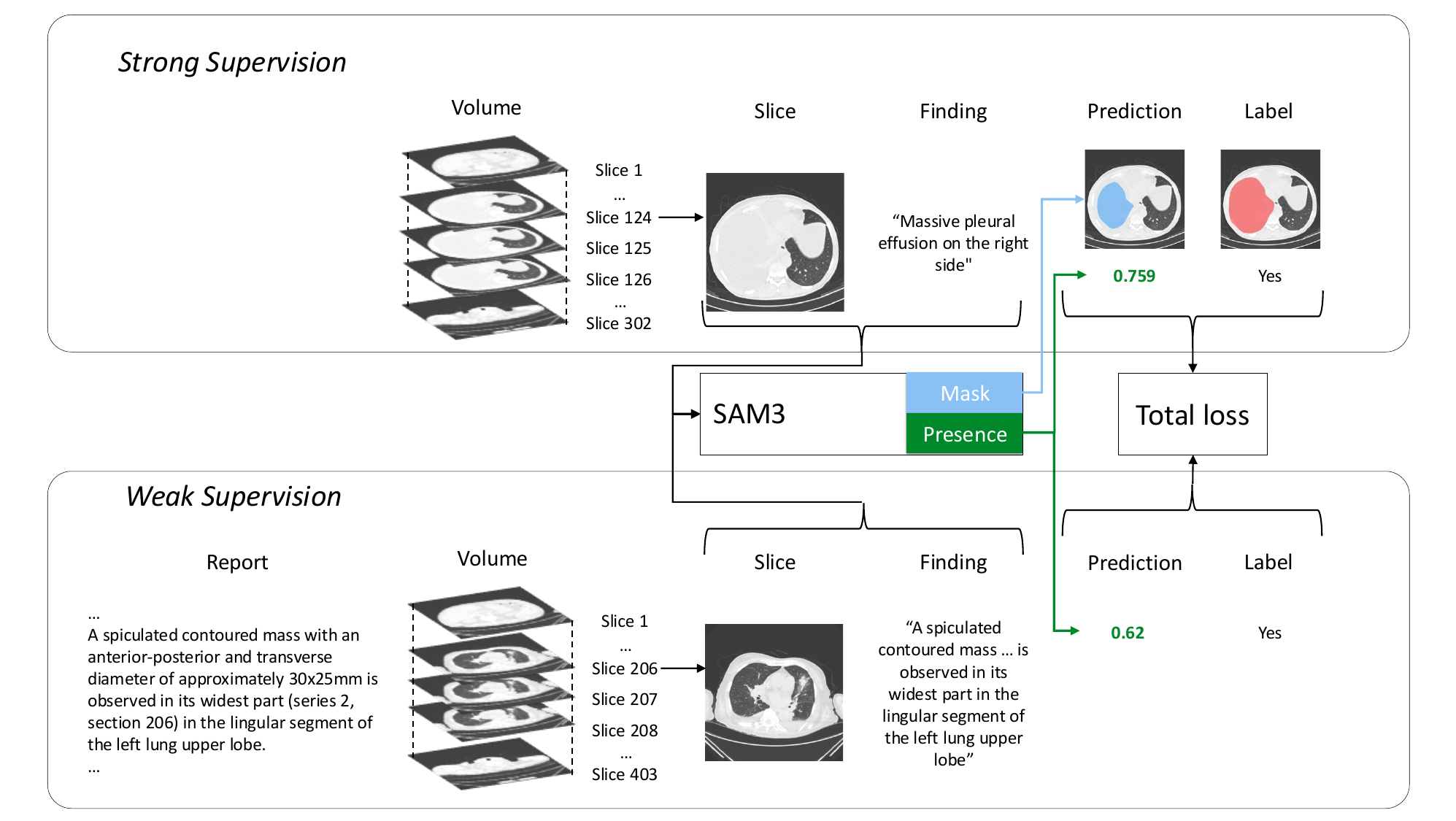}
    \caption{\textbf{Open-ended CT segmentation with weak supervision.} In our method, we train a per-slice segmentation model with both strong and weak supervision. The model takes as input a single 2D slice and a finding description and produces as output a presence (i.e. slice-level existence) probability and segmentation mask. Strongly labeled (volume, finding) pairs include both presence labels and mask labels for every slice but are scarce. Weakly labeled (volume, finding) pairs--which we create from (volume, report) pairs--include only presence labels for certain slices but are widely available. We finetune SAM3 with a loss that combines both sources of data. The examples and predictions in this figure are for illustrative purposes and are not necessarily examples in the training data.}
    \label{fig:method}
\end{figure*}
In this section, we define the language-conditioned CT scan segmentation task, and we describe our approach to it. In particular, we describe (1) the model architecture that we use (SAM3), (2) the procedure that we use for extracting supervision from reports when no ground-truth segmentation is available, and (3) the training recipe. Figure~\ref{fig:method} shows an overview of the method.

\subsection{Task Definition}
\label{sec:task}
In open-ended CT scan segmentation, we are given a CT volume $V: \mathbb{R}^{H \times W \times D}$ (where $H$, $W$, and $D$ denote height, width, and depth) and a textual description $T$ of a finding. We refer to a 2D image obtained by indexing along the depth dimension of $V$ as an image or a slice. The desired output in this task is a segmentation $S: \mathbb{R}^{H \times W \times D}$, in which each value is either $0$ or $1$.
We refer to a training example $(T, V, S)$ that has all three of these elements as a strongly labeled example (or strong supervision). We also consider training examples of the form $(T, V, J)$, where $J \subseteq \{1, 2, ..., D\}$ is a subset of the slice indices such that the ground-truth segmentation mask $S$ is nonzero on the slices in $J$ (and possibly on other slices as well). In this work, we focus on lung findings in Chest CT scans, but our method does not make anatomy-specific assumptions.

\subsection{SAM3 Model}
\label{sec:sam3}
SAM3 \citep{sam3} (part of the Segment Anything model family) takes as input a text description of an entity and a 2D RGB image and returns a binary prediction of whether the entity occurs in the image as well as predictions of where in the image the entity occurs. The components of the SAM3 model that we use are: a text encoder, an image encoder, a fusion transformer, and the segmentation decoder.
The decoder predicts a fixed number of instances, accompanying scores that indicate whether each prediction is active, and an overall score indicating whether the described entity occurs anywhere in the input image. Concretely, the final output of the model, which is produced by the decoder and heads which operate on the decoder output, consists of a presence logit (a scalar), predicted logits $\ell \in \mathbb{R}^{Q \times 1}$, predicted boxes $b \in \mathbb{R}^{Q \times 4}$, and predicted masks $Q \times H \times W$. Here $Q$ is the fixed number of predicted decoder instances, and $H$ and $W$ are the image height and width, respectively. We form the predicted segmentation $\hat{S}$ from these outputs by doing the following for each slice in the input volume: We start from a slice mask initialized to all zeros. If the slice's presence probability is at least $0.5$, then we collect the predicted masks for which the corresponding predicted logit gives an instance probability of at least $0.5$. We then update the slice mask using this collection of predicted masks.

\subsection{Extracting Supervision from Reports}
\label{sec:extract}

The core contribution of our method is that we leverage augment ground-truth segmentation supervision with supervision from language. In order to produce supervision from language, we use an internal database of roughly 244,000 Chest CT reports, which are paired with scans. Each report may be paired with multiple scans (each of which is also called a ``series'').
\begin{algorithm}[htbp]
\caption{Extraction of Supervision Triplets from CT Reports}
\label{alg:supervision_extraction}
\begin{algorithmic}[1]
\Require Database of Chest CT reports $D$, Report Metadata $M$
\Ensure Set of validated supervision triplets $T_{final} = \{(\text{finding}, \text{volume}, \text{slice\_set})\}$
\State $T_{cand} \gets \emptyset$ \Comment{Initialize candidate set}
\For{each report $r \in D$}
    \State $S \gets \text{NormalizeAndRegexLocators}(r)$ \Comment{e.g., ``series X image Y''}
    \State $S \gets \text{SplitLocatorLocalSnippets}(S)$
    \State $S \gets \{s \in S \mid \text{length}(s \text{ without locator}) \ge 8\}$
    \For{each snippet $s \in S$}
        \State $series \gets \text{MatchToSeries}(s, M)$ \Comment{Prefers axial, lung mentions, large slice count}
        \If{$series$ has available RAVE cache}
            \State $slice\_idx \gets \text{MapToTargetSpacing}(s.\text{raw\_idx}, series)$
            \If{$slice\_idx$ is not out of bounds}
                \State $T_{cand} \gets T_{cand} \cup \{(s, series, slice\_idx)\}$
            \EndIf
        \EndIf
    \EndFor
\EndFor
\State $T_{llm} \gets \emptyset$
\For{each candidate $t \in T_{cand}$}
    \State $JSON \gets \text{GPT-5.4-mini}(t.\text{snippet}, \text{extract standalone lung findings and locators})$
    \State $T_{llm} \gets T_{llm} \cup \text{ParseValidLungFindings}(JSON)$
\EndFor
\State $T_{final} \gets \text{RemoveDuplicates}(T_{llm})$ \Comment{Matches study, series, image, and description}
\State $T_{final} \gets \text{CollapseIntervals}(T_{final})$ \Comment{Resolves explicit intervals to representative rows}
\State $T_{final} \gets \text{ApplyAxialFilter}(T_{final})$ \Comment{Discards MIP, COR, SAG oriented series}
\State \Return $T_{final}$
\end{algorithmic}
\end{algorithm}
We use the procedure in Algorithm~\ref{alg:supervision_extraction} to extract triplets of the form $(\textrm{finding description}, \textrm{volume}, \textrm{slice set})$ from reports. See Appendix~\ref{sec:detailed-extraction} for a more detailed description of this procedure. Box~\ref{kf:box:prompt1} provides the prompt given to the language model in this procedure.

At the end of this process, we are left with $93,092$ triplets. Among these, the slice set includes an interval of slices in $2,080$ triplets. Note that the slice set in the triplet may not be exhaustive; it is very common that a single slice is mentioned in the report (e.g. the slice where the finding is most clearly visible) but the finding is visible on other slices as well.

\subsection{Training}
\label{sec:training}

\paragraph{Losses}
We adopt the standard SAM3 loss for finetuning on strongly labeled slices. This loss consists of a presence component, which is a function of the model's per-slice presence prediction, and an instance component, which deals with the model's instance predictions. We describe the presence loss here and provide the remaining details of the loss in Appendix~\ref{sec:full-loss}.

Let $\hat{z}_{c}$ denote the presence logit for slice index $c$, and let $y_c$ denote the ground-truth label for the slice. We set $y_c=1$
if there is a ground-truth instance on the slice and $y_c=0$ otherwise. $\alpha_p$ and $\gamma_p$ are hyperparameters of the focal loss function (see Appendix~\ref{sec:full-loss} for the values that we use). Let $\alpha_t = \alpha_p \cdot y_c + (1 - \alpha_p)(1 - y_c)$. Then the presence loss for slice $c$ is defined as
\begingroup
\small
\begin{align*}
    \mathcal{L}_{pres}(\hat{z}_c) =&\; -\alpha_t \cdot \Big[1 - \left( y_c \sigma(\hat{z}_c) + (1-y_c)(1-\sigma(\hat{z}_c))\right)\Big]^{\gamma_p} \cdot \Big[ y_c \log \sigma(\hat{z}_{c}) + (1 - y_c) \log(1 - \sigma(\hat{z}_{c})) \Big]
\end{align*}
\endgroup

For weakly labeled slices, we apply only the presence loss. The presence loss is defined in the same way for weakly labeled slices as it is for strongly labeled slices, except for a difference in the binary label $y$. For a given weakly labeled example, let $P$ denote the set of indices of slices that are annotated as positive. Recall that $P$ is not necessarily comprehensive; that is, the ground-truth segmentation $S$ may be nonzero on slices whose indices are not in $P$. However, within a single volume, a particular finding typically occurs only on a small number of contiguous slice segments (often only one). Therefore, we set the label $y$ such that it is positive for slice indices in $P$ (and for neighbors thereof, if
only singleton slices, rather than slice intervals, are annotated
) and is negative far away from $P$. Let $P'$ be equal to $P$ if the annotations for the finding include an interval of slices (not only singleton slices); otherwise, let $P' = P \cup \{i: \exists j \in P \text{ and } |i-j| = 1\}$. Concretely, for slice index $c$, we set the label $y_c$ as follows:
\begin{align*}
    y_c =&\; \begin{cases}
        1 & \text{if } c \in P' \\
        0 & \text{if } \{c-g, c-g+1, ..., c+g-1, c+g\} \cap P' = \emptyset \\
        \text{null} & \text{otherwise}
    \end{cases}
\end{align*}
$g$ is a margin hyperparameter that we set to $5$. Only slices with a non-null label participate in the presence loss; let $C$ be the set of all non-null slices in a device's batch. The overall presence loss is $\mathcal{L}_{pres}(\hat{z}) = \frac{1}{|C|}\sum_{c \in C} \mathcal{L}_{pres}(\hat{z}_c)$.

\paragraph{Batch Sampling} For each volume sampled in the batch, we sample a subset of slices. Specifically, we first choose an anchor slice; we choose a positive anchor slice (i.e. a slice which is positive for the finding in question) with probability $0.5$ and otherwise choose a negative anchor slice. Then we form a window of slices with a fixed length (the slice subset size) around the anchor slice--if there are enough slices on either side, then we make the anchor the center of the window; otherwise, we take as many slices as possible on the shorter side and take the remaining on the other side.
\section{Experiments}

\subsection{Evaluation Metrics}
We report two different metrics
\begin{enumerate}
    \item Dice score: equal to the number of voxels  that are positive in both the ground-truth and predicted masks divided by the sum of the number of voxels that are positive in the ground-truth and the number of voxels that are positive in the prediction. In order to measure segmentation performance independent of the performance of presence prediction (slice-wise classification), we also compute a version of dice in which we assume a perfect presence predictor. In this version, the metric is computed based only on slices where the ground-truth mask is non-empty and the predicted masks are not gated by the presence scores.
    \item ROC-AUC: equal to the fraction of (ground-truth positive slice, ground-truth negative slice) pairs such that the positive slice has a greater presence score than the negative slice. Note that this metric evaluates only the model's presence score.
\end{enumerate}

\subsection{Dataset}
For both strong supervision and evaluation, we require a dataset that has finding descriptions, volumes, and associated ground-truth segmentations. To this end, we use the ReXGroundingCT dataset \citep{rexgroundingct}. This dataset is built on the CT-RATE dataset \citep{ctrate}, which contains roughly 25,000 chest CT volumes. The ReXGroundingCT dataset augments CT-RATE with finding annotations and corresponding segmentation masks for 3,192 volumes. For most of the experiments, we use 2830 of the volumes and split the dataset so that there are 1000 volumes in train (corresponding to 2553 findings), 915 volumes in val (2340 findings), and 915 volumes in test (2353 findings). For some experiments, we further subsample the training set to 250 volumes (618 findings) or 500 volumes (1320 findings).

\subsection{Implementation Details}
\paragraph{Data Processing} Unless otherwise specified, we process CT volumes using the RAVE utility \cite{pillar0}. In particular, we resample the volumes to have a spacing of 1.25 mm in each dimension. We then take a center crop of 256 for each slice, and we take a center crop of length 512 along the depth dimension if necessary.
\begin{figure*}[tbp]
    \includegraphics[width=\textwidth]{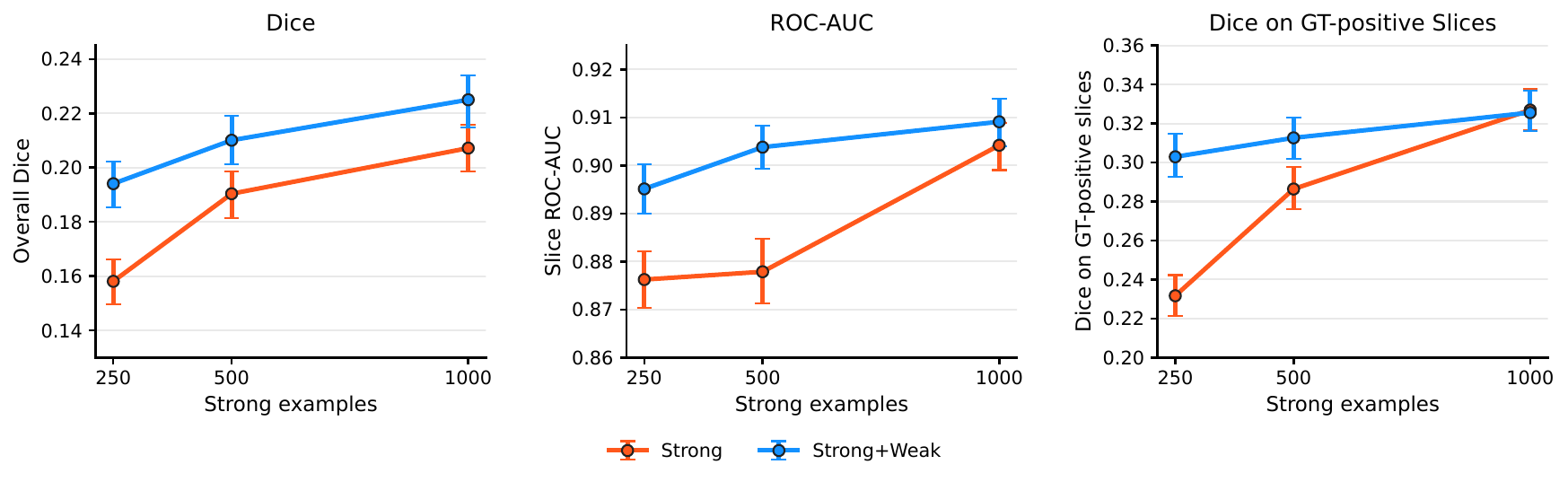}
    \caption{\textbf{Strong vs. Strong+Weak supervision.} ReXGroundingCT test set results with different numbers of strongly labeled volumes. Weak supervision improves segmentation performance, especially when the number of available strongly labeled volumes is modest.}
    \label{fig:three-panel}
\end{figure*}
\begin{figure*}[tbp]
    \includegraphics[width=0.8\textwidth]{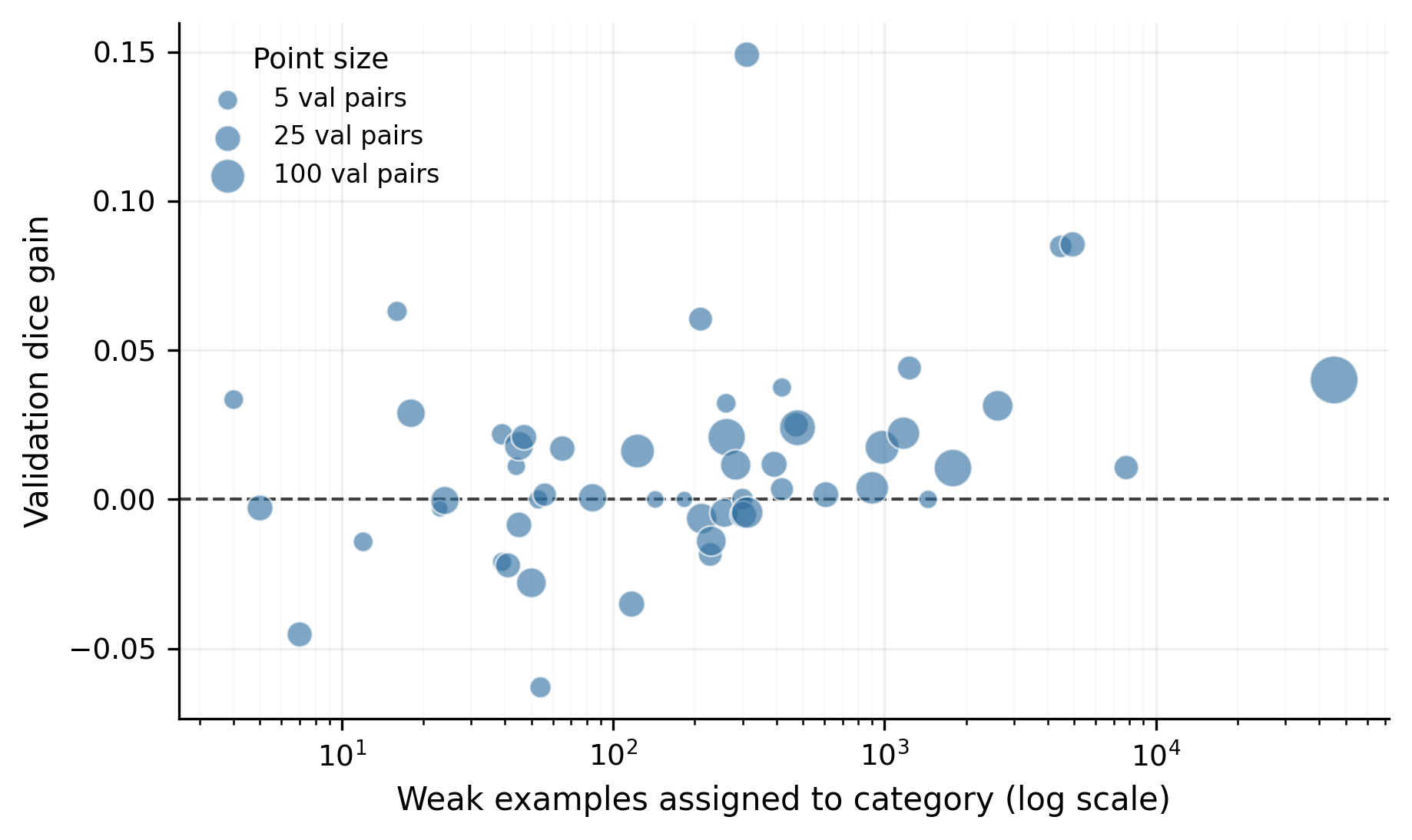}
    \caption{\textbf{Per-category results.} Finding categories with more representation in the weak training data tend to exhibit greater improvement from weakly supervised training. These results are from training on 500 strongly labeled volumes with and without the weak supervision.}
    \label{fig:weak-amount}
\end{figure*}
\paragraph{Hyperparameters} We use learning rates for the transformer, the vision backbone, and the language backbone that have the ratio 16:5:1, respectively, which matches the settings in the recommended SAM3 finetuning settings. Specifically, unless specified otherwise, the learning rates are: $1e-4$ for the vision backbone, $2e-5$ for the language backbone, and $3.2e-4$ for all other parameters. We use the recommended learning rate scheduler, which does linear warmup for 20 steps followed by inverse square root decay. For both the baseline and the weakly supervised method, we experiment with scheduler timescales of 20 and 320 and find that 320 results in the best validation dice in both cases. For the baseline, we experiment with batch sizes of 1 and 5 and find that 1 resulted in the best validation dice.\footnote{The baseline experiments with 500 and 250 volumes encountered NaN loss, so we also ran these experiments (a) with batch size 5 and (b) with batch size 1 and halved learning rates ($5e-5$, $1e-5$, and $1.6e-4$ for the vision backbone, language backbone, and other parameters, respectively); setting (b) yielded the best results (though NaN loss still occurred). For runs in which NaN loss occurred, we attempted to restart from the latest checkpoint.} For the weakly supervised method, we sample 1 strongly labeled instance and 4 weakly labeled instances per batch. These are the other hyperparameters that we use: resize input to $504 \times 504$ before feeding it to the model, $10,000$ training steps, checkpoint/evaluate every $500$ steps
, positive slice anchor probability is $0.5$, presence loss alpha is $0.7$. For both methods, we selected hyperparameters based on validation set results from models trained on the full 1000-volume training set; for experiments on smaller training sets, we used the same hyperparameters.

All models that we train are initialized from the official SAM3 checkpoint \citep{sam3}.

Where we report error bars, they represent 95\% confidence intervals that are computed via the bootstrap with 1000 iterations.

\subsection{Results}
Figure~\ref{fig:three-panel} compares the performance of SAM3 when finetuned with only strong supervision with SAM3 when finetuned with strong+weak supervision at three different levels of strong supervision. The results show that weak supervision improves the performance of the model, especially when the number of strongly labeled volumes is small.
\subsubsection{Per-Category Analysis}
\begin{figure*}[tbp]
    \centering
    \begin{subfigure}{0.8\textwidth}
        \centering
        \includegraphics[width=\linewidth]{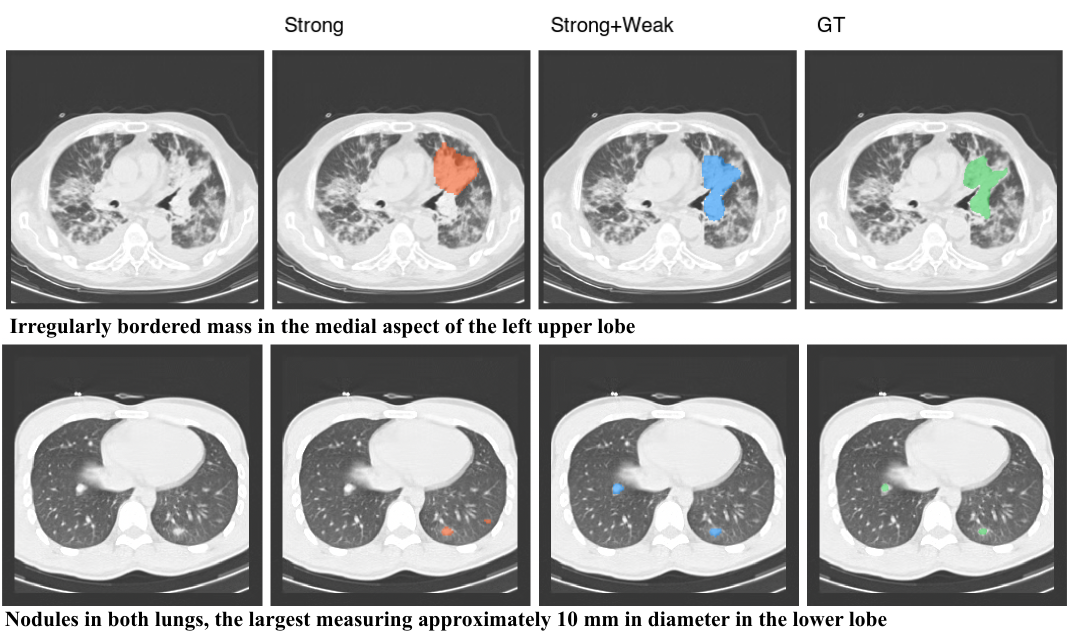}
        \caption{}
        \label{subfig:weak-better}
    \end{subfigure}
    \begin{subfigure}{0.8\textwidth}
        \centering
        \includegraphics[width=\linewidth]{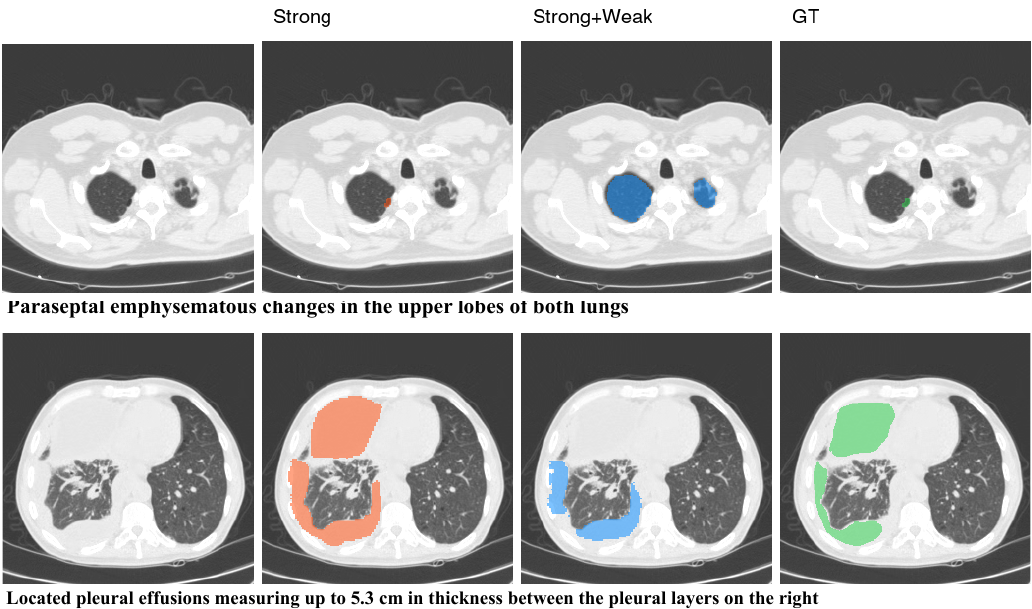}
        \caption{}
        \label{subfig:strong-better}
    \end{subfigure}
    \caption{\textbf{Examples from different categories.} Each 4-image panel shows (from left to right): a slice from the CT volume, the prediction of the strong model, the prediction of the strong+weak model, the ground-truth mask. For each panel, the finding text corresponding to the example appears below the panel. Subfigure~\ref{subfig:weak-better} shows examples from two categories where the strong+weak model is better on average: ``pulmonary mass or mass-like lesion'' and ``solid or nonspecific pulmonary nodules.'' Subfigure~\ref{subfig:strong-better} shows examples from two categories where the strong model is better on average: ``paraseptal emphysema'' and ``large or loculated pleural effusion.''}
    \label{fig:examples}
\end{figure*}
Next, we analyze how the performance difference due to weak supervision varies across categories or finding types and how this variation relates to the amount of weakly labeled data for each category. We prompt a language model (GPT 5.5) to produce a list of categories from all of the train, validation, and test findings in the ReX Grounding dataset (5,848 unique finding texts). For reference, we provide the system prompt in Box~\ref{kf:box:promptcategories} and the user prompt in Box~\ref{kf:box:userprompttaxonomy}.
The resulting set of categories is provided in Table~\ref{tab:ct-finding-taxonomy}. Then we ran language model calls (with GPT 5.5) to assign findings to categories. The system prompt is the same as the one above, and the user prompt is provided in Box~\ref{kf:box:userpromptassign}. Figure~\ref{fig:weak-amount} shows that categories that have greater representation in the weak training data generally exhibit larger gains in dice score from training with the weak supervision. Figure~\ref{fig:examples} shows examples from a few categories with predictions from the strong and strong+weak methods alongside ground-truth segmentations.

\subsubsection{Comparison with State-of-the-art}
\begin{figure*}[tbp]
    \includegraphics[width=\textwidth]{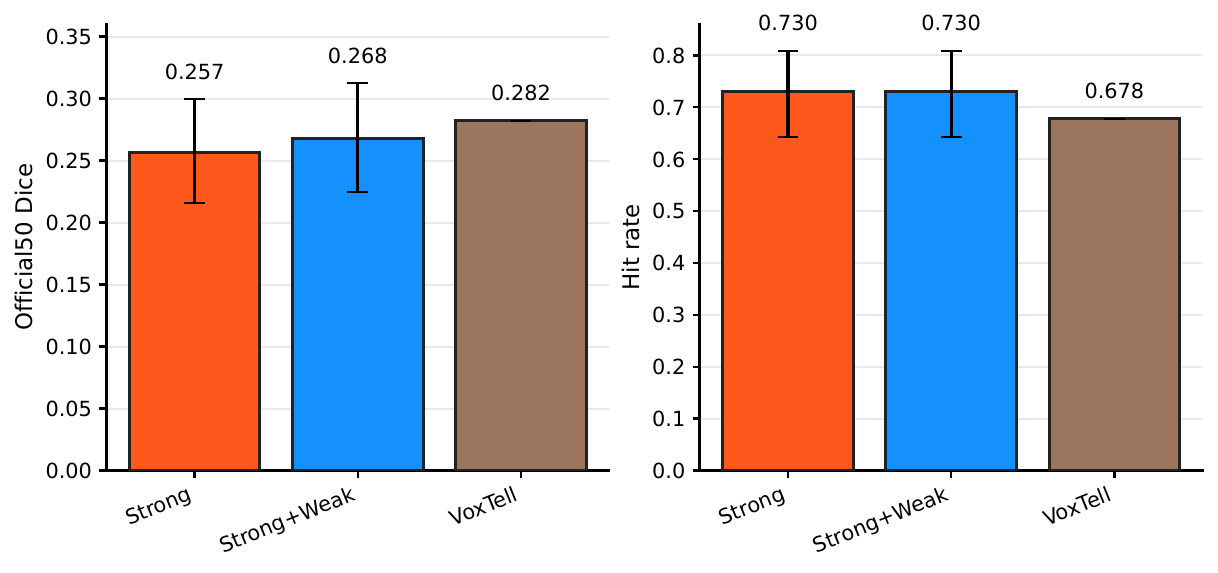}
    \caption{\textbf{Comparison with state-of-the-art.} Our SAM3 models have dice scores which are slightly lower than and hit rates which are slightly greater than those of the state-of-the-art VoxTell model \citep{voxtell}. For this experiment, we train on the full official training set of ReXGroundingCT, except we use a validation set of 100 volumes for model selection. Hit rate is defined as the percent of instances, i.e. (finding, volume) pairs, for which the dice score is at least 0.05.}
    \label{fig:sota}
\end{figure*}
\begin{figure*}[tbp]
    \centering
    \includegraphics[width=0.6\textwidth]{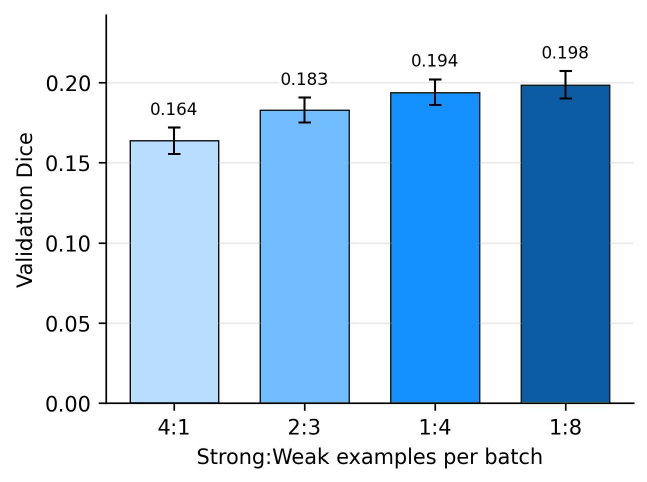}
    \caption{\textbf{Ablating strong:weak ratio.} Adding more weak supervision per batch improves the dice score, though there are diminishing returns.}
    \label{fig:ablation}
\end{figure*}
Figure~\ref{fig:sota} compares the performance of our strong and strong+weak finetuned SAM3 models with that of the state-of-the-art VoxTell model \citep{voxtell} using the official train/val split of RexGroundingCT. In this split, we use 2892 volumes for training. Both finetuned SAM3 models have slightly lower dice and slightly greater hit rate than the VoxTell model, but the differences do not exceed the width of the error bars. The strong+weak model has slightly higher dice score than the strong model in this full-training data setting. For this experiment, we train and evaluate SAM3 models with the images from the raw volumes rather than RAVE-cached versions of the volumes.

\subsubsection{Ratio of strong to weak supervision.}
One of the main design choices specific to the weak supervision method is the ratio of strongly labeled to weakly labeled instances in a single batch. Figure~\ref{fig:ablation} presents the results of an ablation study for this hyperparameter. We find that decreasing the strong:weak ratio generally improves results, though there are diminishing returns.

\section{Conclusion}
In this work, we present a method for extracting and using scalable weak supervision from radiology reports for the task of open-ended CT volume segmentation. Specifically, we extract from reports finding descriptions with indices of slices on which those findings are visible. We finetune a 2D image segmentation model on a mix of strongly labeled data (with ground-truth segmentation masks) and weakly labeled data (with only slice-level binary labels for a subset of slices). Our experiments show that when strongly labeled data is scarce, the weakly labeled data improves segmentation quality. These results raise several research questions for future work: How well does this method work on other types of CT scans and other medical imaging modalities? How can this source of weak supervision be used to train a 3D-native segmentation model? What other objectives and/or forms of supervision from reports could further improve performance?

\bibliographystyle{plainnat}
\bibliography{references}


\appendix
\clearpage
\section{Appendix}

\subsection{Extraction of Weak Supervision from Reports}
\label{sec:detailed-extraction}
We apply the following procedure to extract weak supervision from reports.
\begin{enumerate}
      \item Normalize report text and split it into candidate sentence-like snippets. Use regular expressions to identify snippets containing
  explicit series/image locators, including patterns such as ``series X image Y'' and image intervals such as ``series X images Y--Z''.
      \item For snippets with multiple locator mentions, split them into locator-local candidate snippets when possible.
      \item Discard candidates whose text is shorter than 8 characters after removing the locator text.
      \item Use the referenced series number to match the candidate to a series in the report metadata, requiring an available RAVE cache for that
  series. If multiple series match, prefer series whose descriptions are not MIP/SAG/COR/etc., are axial-like, mention lung, and have larger slice
  count.
      \item Compute the conversion factor from raw slice indices in the series to slice indices in the resampled volume. (We resample volumes to a fixed target spacing before passing them to the model) When available, this uses the ratio between
  original and target z-spacing; otherwise it falls back to a shape-ratio mapping. Candidates whose source image index is out of range are discarded.
      \item Use GPT-5.4-mini to postprocess the regex-prefiltered report snippets. The model returns strict JSON containing one or more candidate lung
  findings, each with a standalone description and one or more explicit series/image locators. For reference, we provide the prompt in Box~\ref{kf:box:prompt1} . Retain only lung findings with valid locators that can be
  matched to a series and mapped to a RAVE slice.
      \item Remove duplicate rows, using the study, series, image number, and processed description/locator identity.
      \item Detect explicit image intervals in the raw text, attach interval metadata, and collapse expanded interval rows to a single representative
  row.
  metadata.
      \item Finally, apply an axial-only filter: drop MIP series, drop COR/SAG series descriptions that do not also indicate axial acquisition, and
  drop rows whose RAVE metadata orientation is coronal or sagittal.
\end{enumerate}

\subsection{SAM3 Loss}
\label{sec:full-loss}
Let $Q$ denote the number of predicted object queries in each query branch of the DETR model (there is a one-to-one, o2o, branch and a one-to-many, o2m, branch), let $\hat{z}$ denote the presence logits, let $\hat{p}_i$ denote the predicted confidence probability for a particular prediction $i$, let $\hat{b_i}$ denote the predicted box for a matched prediction $i$, let $b_i$ denote the ground-truth box for a matched prediction $i$, let $\hat{m_i}$ denote the predicted per-pixel mask logits for a matched prediction $i$ (flattened to a vector), let $m_i$ denote the ground-truth binary per-pixel mask for a matched prediction $i$ (flattened to a vector), let $B_{\text{device}}$ denote the number of slices in a device's batch, and let $\sigma(\cdot)$ denote the sigmoid function.

These are the hyperparameters involved in the loss, along with their values in our experiments (unless otherwise specified): $Q=200$, $\alpha_{p}=0.7$, $\gamma_p=0$, $\alpha_{\text{ce}}=0.25$, $w_{\text{pos}}=20$, $\gamma_{\text{ce}}=2$, $\alpha_{\text{mask}}=0.25$, $\gamma_{\text{mask}}=2$, $w_{\text{pres}}=20$, $w_{\text{ce}}=20$, $w_{L1}=5$, $w_{GIoU}=2$, $w_{\text{MaskFocal}}=5$, $w_{\text{dice}}=20$, and $\lambda_{o2m}=2$.

The standard SAM3 losses include a slice-wise presence loss and instance-wise losses.

The presence loss $\mathcal{L}_{pres}(\hat{z})$ is defined in Section~\ref{sec:training}.

Next, we define the per-instance classification loss.  For a matched prediction $i$, we define a soft target $t_i = \text{stopgrad}\left[\text{clamp}\!\Big(\hat{p}_i^{\;\alpha_{\text{ce}}} \cdot \text{IoU}(\hat{b}_i^{xyxy},\; b_i^{xyxy})^{1-\alpha_{\text{ce}}},\;\; 0.01,\;\; 1\Big)\right]$. For a matching $M$ between (some subset of) prediction queries and ground-truth instances, we define a positive loss $\mathcal{L}_{\text{ce}}^{+}(\hat{p}, \hat{b}, M) = -\frac{w_{\text{pos}}}{Q} \sum_{i \in M} \Big[ t_i \log \hat{p}_i + (1 - t_i) \log(1 - \hat{p}_i) \Big]$ and a negative loss $\mathcal{L}_{\text{ce}}^{-}(\hat{p}, M) = -\frac{1}{Q} \sum_{j \notin M} \hat{p}_j^{\;\gamma_{\text{ce}}} \cdot \log(1 - \hat{p}_j)$. Then the combined per-instance classification loss is defined as $\mathcal{L}_{\text{ce}}(\hat{p}, \hat{b}, M) = \frac{1}{B_{\text{device}}}\left[\mathcal{L}_{\text{ce}}^{+}(\hat{p}, \hat{b}, M) + \mathcal{L}_{\text{ce}}^{-}(\hat{p}, M)\right]$.

The bounding box loss is a combination of two terms: an L1 loss and a generalized IoU (GIoU) loss. Both losses are applied to matched predictions. Let $N_{gt}$ be the number of ground-truth instances in each device's local batch, averaged across devices and clamped such that it is at least $1$. $\mathcal{L}_{L1}(\hat{b}, M) = \frac{1}{N_{gt}} \sum_{i \in M} \left| \hat{b}_i - b_i \right|_1$. $\mathcal{L}_{GIoU}(\hat{b}, M) = \frac{1}{N_{gt}} \sum_{i \in M} \Big(1 - \text{GIoU}(\hat{b}_i^{xyxy},\; b_i^{xyxy})\Big)$, where $\text{GIoU}(A, B) = \text{IoU}(A, B) - \frac{|\text{MinimalEnclosingBox}(A, B) \setminus (A \cup B)|}{|\text{MinimalEnclosingBox}(A, B)|}$.

The mask loss is a combination of a per-pixel sigmoid focal loss and a dice loss, both applied to matched predictions. For the focal loss, for a matched prediction $i$ and pixel $k$, let $p(i,k) = \sigma(\hat{m}_{ik}) \cdot m_{ik} + (1 - \sigma(\hat{m}_{ik}))(1 - m_{ik})$, and $w(i,k) = \alpha_{\text{mask}} \cdot m_{ik} + (1-\alpha_{\text{mask}})(1-m_{ik})$. Then the mask focal loss is defined as $\mathcal{L}_{\text{MaskFocal}}(\hat{m}, M) = \frac{1}{N_{gt}} \sum_{i \in M} \frac{1}{H \cdot W} \sum_{k=1}^{H \cdot W} -w(i,k) \cdot (1 - p(i,k))^\gamma_{\text{mask}} \cdot \left(\log(\sigma(\hat{m}_{ik})) \cdot m_{ik} + \log(1 - \sigma(\hat{m}_{ik}))(1 - m_{ik})\right)$. The dice loss is defined as $\mathcal{L}_{\text{dice}} = \frac{1}{N_{gt}} \sum_{i \in M} \left(1 - \frac{2 \sum_k \sigma(\hat{m}_{ik}) \cdot m_{ik} + 1}{\sum_k \sigma(\hat{m}_{ik}) + \sum_k m_{ik} + 1}\right)$.

These losses are applied on the outputs of each decoder layer.  For layers $l$ before the final layer, the instance loss is given by $\mathcal{L}^{l}(M) = w_{\text{ce}} \cdot \mathcal{L}_{\text{ce}}(\hat{p}^l, \hat{b}^l, M) + w_{L1} \cdot \mathcal{L}_{L1}(\hat{b}^l, M) + w_{GIoU} \cdot                   \mathcal{L}_{GIoU}(\hat{b}^l, M)$, where we use the $l$ superscript to denote predictions in layer $l$. For the final layer $l_{\text{final}}$, $\mathcal{L}^{l_{\text{final}}}(\hat{p}^{l_{\text{final}}}, \hat{b}^{l_{\text{final}}}, \hat{m}, M) = w_{\text{ce}} \cdot \mathcal{L}_{\text{ce}}(\hat{p}^{l_{\text{final}}}, \hat{b}^{l_{\text{final}}}, M) + w_{L1} \cdot \mathcal{L}_{L1}(\hat{b}^{l_{\text{final}}}, M) + w_{GIoU} \cdot                   \mathcal{L}_{GIoU}(\hat{b}^{l_{\text{final}}}, M) + w_{\text{MaskFocal}} \cdot \mathcal{L}_{\text{MaskFocal}}(\hat{m}^{l_{\text{final}}}, M) + w_{\text{dice}} \cdot \mathcal{L}_{\text{dice}}(\hat{m}^{l_{\text{final}}}, M)$.
The overall loss is the sum of the per-layer losses across both the one-to-one (o2o) and one-to-many (o2m) branches:
\begin{align*}
\mathcal{L} =&\; \sqrt{B_{\text{device}}}\Big[w_{pres} \mathcal{L}_{pres}(\hat{z}^{l_{\text{final}}})+\mathcal{L}^l(M_{o2o,l_{\text{final}}}) + \lambda_{o2m} \mathcal{L}^l(M_{o2m,l_{\text{final}}}) \\
&\;+\sum_{l=1}^{l_{\text{final}}-1} w_{pres} \mathcal{L}_{pres}(\hat{z}^l)+\mathcal{L}^l(M_{o2o,l}) + \lambda_{o2m} \mathcal{L}^l(M_{o2m-aux,l})\Big]
\end{align*}
, where $M_{o2o,l}$ denotes a one-to-one matching computed on the o2o branch predictions in layer $l$, $M_{o2m,l_{\text{final}}}$ denotes a one-to-many matching on the o2m branch predictions in layer $l$, and $M_{o2m-aux,l}$ denotes a one-to-one matching computed on the o2m branch predictions in layer $l$. Please see \citet{sam3} for an explanation of these matchings.

\subsection{Prompts}
Boxes~\ref{kf:box:prompt1}, \ref{kf:box:promptcategories}, \ref{kf:box:userprompttaxonomy}, and \ref{kf:box:userpromptassign} provide the prompts we used in various steps.
\begin{keyfinding}{System prompt for filtering and processing finding texts}{box:prompt1}
You extract localized lung findings from chest CT report snippets.

Return only strict JSON with this schema:
\begin{quote}\ttfamily
\{\\
\hspace*{1em}"findings": [\\
\hspace*{2em}\{\\
\hspace*{3em}"is\_lung\_finding": true,\\
\hspace*{3em}"standalone\_description": "one self-contained finding sentence with no series/image references",\\
\hspace*{3em}"locators": [\\
\hspace*{4em}\{\\
\hspace*{5em}"series\_number": "3",\\
\hspace*{5em}"image\_number": 51,\\
\hspace*{5em}"image\_number\_end": null,\\
\hspace*{5em}"evidence": "short quote tying this locator to this finding"\\
\hspace*{4em}\}\\
\hspace*{3em}]\\
\hspace*{2em}\}\\
\hspace*{1em}]\\
\}
\end{quote}

Rules:

\hangindent=1.5em \makebox[1.5em][l]{-} Return one object per distinct lung finding in the raw text.

\hangindent=1.5em \makebox[1.5em][l]{-} If a snippet contains multiple findings, split them. Do not merge several findings into one description.

\hangindent=1.5em \makebox[1.5em][l]{-} If different anatomical locations or abnormalities have different image locators, return separate finding objects with different \texttt{standalone\_description} values.

\hangindent=1.5em \makebox[1.5em][l]{-} Use multiple locators in one finding only when the same finding is shown at multiple discrete images or image intervals.

\hangindent=1.5em \makebox[1.5em][l]{-} A lung finding includes lung nodules, pulmonary nodules/masses/opacities, consolidation, ground-glass opacity, emphysema, atelectasis, pleural effusion, pneumothorax, airway findings, and other pulmonary or pleural abnormalities.

\hangindent=1.5em \makebox[1.5em][l]{-} The \texttt{standalone\_description} must be understandable without other report context and must not mention "series", "image", "as above", "this", "same", or other report/image references.

\hangindent=1.5em \makebox[1.5em][l]{-} Each locator must contain one series number and one image number. If the raw text says "(series 5 image 21) ... (image 28)", use \texttt{series\_number} "5" for both locators.

\hangindent=1.5em \makebox[1.5em][l]{-} If a locator is a true interval such as "images 10-14" or "images 10 through 14", set \texttt{image\_number} to 10 and \texttt{image\_number\_end} to 14.

\hangindent=1.5em \makebox[1.5em][l]{-} Do not convert comma-separated or conjunction-separated image mentions into intervals. "images 21, 28" means two locators: 21 and 28, both with \texttt{image\_number\_end} null.

\hangindent=1.5em \makebox[1.5em][l]{-} If no explicit series/image locator is tied to the finding, use an empty locators list.
\end{keyfinding}

\begin{keyfinding}{System prompt for defining a taxonomy for findings and assigning findings to categories in the taxonomy}{box:promptcategories}
You are helping define a clinically meaningful taxonomy for CT chest finding texts.

The taxonomy will be used to stratify model performance for a slice/mask localization task. Favor categories that are:

\hangindent=1.5em \makebox[1.5em][l]{-} clinically interpretable,

\hangindent=1.5em \makebox[1.5em][l]{-} coarse enough to aggregate many near-duplicate finding descriptions,

\hangindent=1.5em \makebox[1.5em][l]{-} fine enough to separate localization-relevant entities when useful.

Use the finding texts as evidence. Merge wording variants such as "tiny", "small", "subcentimeter", and location-only changes unless the distinction is clinically or visually important. Keep important visual distinctions when the texts support them, for example solid/nonspecific nodules vs ground-glass nodules/opacities, atelectasis vs consolidation, airway wall thickening vs bronchiectasis, and pleural effusion vs pleural thickening.

Return only valid JSON. Do not include markdown fences.
\end{keyfinding}

\begin{keyfinding}{User prompt for defining a taxonomy for findings}{box:userprompttaxonomy}
Propose one final taxonomy for all supplied CT chest finding texts.

Target size: \texttt{\{target\_min\_categories\}} to \texttt{\{target\_max\_categories\}} categories. You must return at least \texttt{\{target\_min\_categories\}} categories. Do not collapse the taxonomy into only nodules or any other single dominant class. For each category, include only a small representative set of finding IDs; do not attempt to assign every finding in this taxonomy response. A later call will perform exhaustive finding-to-category assignment.

The taxonomy will be used to stratify CT chest slice/localization model performance, so do not create categories that are only left/right/lobe/location variants. The categories should be coarse clinical concepts such as nodules, ground-glass opacities, atelectasis, emphysema, airway disease, pleural disease, lymph nodes, consolidation, fibrosis/scarring, etc., but use the supplied texts as the evidence for the final category list.

Return JSON with this schema:
\begin{quote}\ttfamily
\{\\
\hspace*{1em}"taxonomy\_version": "ct\_chest\_findings\_v0",\\
\hspace*{1em}"categories": [\\
\hspace*{2em}\{\\
\hspace*{3em}"id": "lower\_snake\_case\_stable\_id",\\
\hspace*{3em}"name": "Human-readable category name",\\
\hspace*{3em}"parent": "optional broader parent or null",\\
\hspace*{3em}"definition": "What belongs in this category",\\
\hspace*{3em}"include\_patterns": ["phrases or concepts that should map here"],\\
\hspace*{3em}"exclude\_patterns": ["nearby concepts that should not map here"],\\
\hspace*{3em}"example\_finding\_ids": ["finding\_00001", "finding\_00002"]\\
\hspace*{2em}\}\\
\hspace*{1em}],\\
\hspace*{1em}"recommended\_default\_category": "other\_or\_uncertain",\\
\hspace*{1em}"notes": ["brief notes about ambiguous boundaries"]\\
\}
\end{quote}

Finding texts:\\
\texttt{\{compact\_findings\_json\}}
\end{keyfinding}

\begin{keyfinding}{User prompt for assigning a finding to a taxonomy category}{box:userpromptassign}
Assign each finding text to exactly one taxonomy category.

Return JSON with this schema:
\begin{quote}\ttfamily
\{\\
\hspace*{1em}"assignments": [\\
\hspace*{2em}\{\\
\hspace*{3em}"finding\_id": "finding\_00001",\\
\hspace*{3em}"taxonomy\_id": "one\_category\_id\_from\_taxonomy",\\
\hspace*{3em}"confidence": 0.0\\
\hspace*{2em}\}\\
\hspace*{1em}]\\
\}
\end{quote}

Taxonomy:\\
\texttt{\{compact\_taxonomy\_json\}}

Finding texts to assign:\\
\texttt{\{compact\_findings\_json\}}
\end{keyfinding}

\subsection{Categorization of Findings}
Table~\ref{tab:ct-finding-taxonomy} provides the taxonomy used in the category-wise evaluation.
\begin{table*}[tbp]                                                                                                                                                                                                    
  \centering                                                                                                                                                                                                          
  \tiny                                                                                                                                                                                                              
  \caption{CT chest finding taxonomy used for per-category analysis.}                                                                                                                                                 
  \label{tab:ct-finding-taxonomy}                                                                                                                                                                                     
  \begin{tabular}{r l}                                                                                                                                                                                                
  \toprule                                                                                                                                                                                                            
  \# & Category \\                                                                                                                                                                                                    
  \midrule                                                                                                                                                                                                            
  1 & Solid or nonspecific pulmonary nodules \\                                                                                                                                                                       
  2 & Calcified pulmonary nodules or granulomas \\                                                                                                                                                                    
  3 & Ground-glass, subsolid, or semisolid nodules \\                                                                                                                                                                 
  4 & Perifissural or intrapulmonary lymph node-type nodules \\                                                                                                                                                       
  5 & Metastatic pulmonary nodules or lesions \\                                                                                                                                                                      
  6 & Centrilobular, centriacinar, acinar, or tree-in-bud nodules \\                                                                                                                                                  
  7 & Miliary or diffuse micronodular disease \\                                                                                                                                                                      
  8 & Pulmonary or hilar mass / mass-like lesion \\                                                                                                                                                                   
  9 & Cavitary lesions, cavitary nodules, or abscess-like cavities \\
  10 & Air cysts, bullae, and blebs \\
  11 & Emphysema, general or centriacinar/panacinar \\
  12 & Paraseptal or subpleural emphysema \\
  13 & Mosaic attenuation, mosaic perfusion, or air trapping \\
  14 & Ground-glass opacities, general \\
  15 & Patchy, peripheral, or subpleural ground-glass opacities \\
  16 & Diffuse or multilobar ground-glass opacities \\
  17 & Round or nodular ground-glass opacities \\
  18 & Ground-glass with halo sign or vascular enlargement \\
  19 & Crazy paving pattern \\
  20 & Consolidation, general \\
  21 & Pneumonic, viral, COVID-19, or atypical infiltration \\
  22 & Consolidation with air bronchograms \\
  23 & Nodular or round consolidation \\
  24 & Subpleural or peripheral consolidation \\
  25 & Mixed ground-glass opacity and consolidation \\
  26 & Atelectasis, general \\
  27 & Linear, band-like, or subsegmental atelectasis \\
  28 & Dependent, passive, or compressive atelectasis \\
  29 & Lobar collapse or complete/near-complete atelectasis \\
  30 & Fibroatelectatic changes \\
  31 & Peribronchial or bronchial wall thickening \\
  32 & Bronchiectasis \\
  33 & Traction or cicatricial bronchiectasis \\
  34 & Bronchial ectasia or dilatation \\
  35 & Airway luminal obstruction, mucus, or secretions \\
  36 & Tracheal or large airway abnormality \\
  37 & Pleural effusion \\
  38 & Large or loculated pleural effusion \\
  39 & Pneumothorax or hydropneumothorax \\
  40 & Bronchopleural fistula \\
  41 & Pleural thickening, plaques, or pleural calcification \\
  42 & Fissural thickening or fissural abnormality \\
  43 & Pulmonary scarring or fibrosis, general \\
  44 & Apical pleuroparenchymal scarring or fibrosis \\                                                                      
  45 & Fibrotic bands, parenchymal bands, and subpleural lines \\                                                            
  46 & Honeycombing or fibrotic interstitial lung disease \\                                                                 
  47 & Interlobular septal or interstitial thickening \\                                                                     
  48 & Architectural distortion or volume loss \\             
  49 & Dependent density or nonspecific density increase \\                                                                  
  50 & Reticular or reticulonodular opacities/infiltrates \\                                                                 
  51 & Pulmonary vascular enlargement or prominence \\                                                                       
  52 & Pulmonary vascular malformation or sequestration \\                                                                   
  53 & Subcutaneous or chest wall emphysema \\                
  54 & Chest wall or extrapulmonary thoracic mass \\                       
  55 & Air bronchogram sign \\                                             
  56 & Halo or reverse halo sign \\                                        
  57 & Other or uncertain finding \\                                       
  \bottomrule                                                              
  \end{tabular}                                                        
\end{table*}


\end{document}